\documentclass[lettersize,journal]{IEEEtran}
\usepackage{cite}
\usepackage{amsmath,amssymb,amsfonts}
\usepackage{balance}
\usepackage{graphicx,comment}
\usepackage{textcomp}
\usepackage{xcolor}
\usepackage{algorithm}
\usepackage{algpseudocode}
\usepackage{listings}
\usepackage{xcolor}
\usepackage{subcaption}

\def\BibTeX{{\rm B\kern-.05em{\sc i\kern-.025em b}\kern-.08em
    T\kern-.1667em\lower.7ex\hbox{E}\kern-.125emX}}
\begin{document}

\title{A Novel MLLM-based Approach for Autonomous Driving in Different Weather Conditions}

\author{Sonda Fourati, Wael Jaafar,~\IEEEmembership{Senior Member,~IEEE,} and Noura Baccar
	\thanks{S. Fourati and N. Baccar are with the Computer Systems Engineering Department, Mediterranean Institute of Technology (MedTech), Tunis, Tunisia. W. Jaafar is with the Software and IT Engineering Department, Ecole de Technologie Supérieure (ÉTS), Montreal, QC H3C 1K3, Canada. E-mails: \{sonda.fourati, noura.baccar\}@medtech.tn; wael.jaafar@etsmtl.ca.}
	\thanks{Manuscript received XX XX, 20XX; .}}

\markboth{Journal of \LaTeX\ Class Files,~Vol.~14, No.~8, August~20XX}%
{Shell \MakeLowercase{\textit{et al.}}: A Sample Article Using IEEEtran.cls for IEEE Journals}

\IEEEpubid{0000--0000/00\$00.00~\copyright~2021 IEEE}

\maketitle

\begin{abstract}
Autonomous driving (AD) technology promises to revolutionize daily transportation by making it safer, more efficient, and more comfortable. Their role in reducing traffic accidents and improving mobility will be vital to the future of intelligent transportation systems. Autonomous driving in harsh environmental conditions presents significant challenges that demand robust and adaptive solutions and require more investigation. In this context, we present in this paper a comprehensive performance analysis of an autonomous driving agent leveraging the capabilities of a Multi-modal Large Language Model (MLLM) using GPT-4o within the LimSim++ framework that offers close loop interaction with the CARLA driving simulator. We call it MLLM-AD-4o. Our study evaluates the agent's decision-making, perception, and control under adverse conditions, including bad weather, poor visibility, and complex traffic scenarios. Our results demonstrate the AD agent's ability to maintain high levels of safety and efficiency, even in challenging environments, underscoring the potential of GPT-4o to enhance autonomous driving systems (ADS) in any environment condition. Moreover, we evaluate the performance of MLLM-AD-4o when different perception entities are used including either front cameras only, front and rear cameras, and when combined with LiDAR. 
The results of this work provide valuable insights into integrating MLLMs with AD frameworks, paving the way for future advancements in this field.
\end{abstract}


\section{Introduction}
Autonomous driving systems (ADS) are set to play a pivotal role in the future, necessitating advanced technology to ensure their safe and efficient integration into transportation networks. Artificial Intelligence (AI), in particular large language models (LLMs) and multimodal LLMs (MLLMs), can be used to enhance the decision-making capabilities of ADS \cite{chang2024survey}\cite{fourati2024xlm}. 

Typically, MLLMs combine multi-modal data including images,
video, and audio data with the advanced reasoning capabilities of LLMs \cite{wang2024exploring}. Hence, they can act as decision-making agents for various applications including medicine, education, and intelligent transport systems (ITS). More accurate and timely responses to dynamic driving conditions can be achieved when integrated into ADS. Specifically, they can process vast amounts of driving data to understand and predict complex traffic patterns, thus enabling autonomous vehicles (AVs) to learn from their environment and enhance their driving performances. 
In addition, MLLMs contribute significantly to the development of ADS by providing the computational power required to analyze the driving scene in real time and make accurate decisions \cite{cui2024survey}. 

In this context, various approaches have been proposed to use MLLMs for AD. 
Indeed, several strategies have been proposed based on prompt engineering \cite{wen2023dilu,ding2023hilm,yuan2024rag}, fine-tuning \cite{wang2024drivecot,wang2023drivemlm,huangdrivlme,liao2024vlm2scene,wang2024omnidrive}, and reinforcement learning with human feedback (RLHF) \cite{tian2024enhancing,mao2023language,rayfeedback}. Furthermore, MLLMs toward AD provisioning could be utilized for perception and scene understanding  \cite{ashqar2024leveraging,luo2024delving,li2024dense}, question answering \cite{xu2023drivegpt4}, planning and control \cite{wang2023drivemlm,cui2024receive}, and multitasking \cite{li2024unifiedmllm}. 
Despite their diverse methods and applications, MLLMs experience limitations in ADS. For instance, fine-tuning-based approaches are computationally expensive and may result in models overfitting for specific tasks or environments. Also, RLHF can align models with human preferences, however, it still faces scaling and generalization issues beyond particular training data instances. Finally, most works focused on using MLLM for ADS in perfect and clear weather conditions, which does not reflect the reality of many areas around the globe.   

Consequently, in this work, we introduce 
MLLM-AD-4o, a novel prompt engineering MLLM-based approach for AD, and capable of handling autonomous driving in any weather condition. Our approach integrates different sensor modalities, including LiDAR and cameras, allowing the system to adapt according to available sensor data. Based on prompt engineering, MLLM-AD-4o provides enhanced environment perception, scene understanding, reasoning, and decision-making in ADS. 
Hence, our contributions can be summarized as follows: 
\begin{itemize}
 \item Unlike previous works, we integrate for the first time the harsh environment driving conditions into the CARLA simulator by leveraging functions and modules from the LimSim++ framework. This work has been documented and disseminated in Github for open public access and reproducibility \cite{GithubSonda}.      
  \item We propose MLLM-AD-4o, a novel MLLM-based AD agent that leverages GPT-4o \cite{islam2024gpt} for driving perception and control decision-making. 
 \item Through extensive experiments, we conduct a performance analysis of the proposed MLLM-AD-4o in terms of safety, comfort, efficiency, and speed score metrics, and under different weather conditions and types of involved sensor data (e.g., front and/or rear cameras and/or LiDAR).
 \end{itemize}

\newpage
The remainder of the paper is organized as follows. In Section II, we present the related works that integrate LLMs/MLLMs into driving agents in a closed-loop using the CARLA simulator. In Section III, we explain the basic concepts of ADS and LLM/MLLMs with a focus on the GPT-4o architecture and functions. Section IV presents the architecture, main modules, and technical details for integrating an MLLM agent driver in a closed-loop environment using Limsim++. Section V evaluates the performance of MLLM-AD-4o in diverse scenarios and conditions. Finally, Section VI closes the paper.

\section{Related Work}
Recently, there has been a significant effort to enhance the efficiency and reliability of ADS using LLMs and MLLMs. Among proposed driving agents, we focus here on the contributions based on prompt engineering. 
Authors of \cite{chen2023driving} proposed ``LLM-driver'', a framework integrating numeric vector modalities, into pre-trained LLMs for question-answering (QA). LLM-driver uses object-level 2D scene representations to fuse vector data into a pre-trained LLM with adapters. A language generator (lanGen) is used to ground vector representations into LLMs. 
The model’s performance was evaluated on perception and action prediction using mean absolute error (MAE) for predictions, accuracy of traffic light detection, and normalized errors for acceleration, brake pressure, and steering. In \cite{jin2023surrealdriver}, the SurrealDriver framework has been proposed. It consists of an LLM-based generative driving agent simulation framework with multitasking capabilities that integrate perception, decision-making, and control processes to manage complex driving tasks. 
In \cite{yang2024driving}, the authors developed LLM-based driver agents with reasoning and decision-making abilities, aligned with human driving styles. Their framework utilizes demonstrations and feedback to align the agents' behaviors with those of humans, utilizing data from human driving experiments and post-driving interviews. 
Also, the authors of \cite{guo2024co} proposed the Co-driver framework for planning \& control and trajectory prediction tasks. Co-driver utilizes prompt engineering to understand visual inputs and generate driving instructions, while it uses deep reinforcement learning (DRL) for planning and control. 
In \cite{wu2023language}, the authors introduced the PromptTrack framework, an approach to 3D object detection and tracking, by integrating cross-modal features within prompt reasoning. PromptTrack uses language prompts that act as semantic guides to enhance the contextual understanding of a scene. 
Finally, authors in \cite{ding2023hilm} proposed HiLM-D as an efficient technique to incorporate high-resolution information in ADS for hazardous object localization. 

Despite the variety of proposed driving agents, to our knowledge, no study has fully assessed the performance of MLLM-based driving agents within a closed-loop framework under harsh environmental conditions. Table~\ref{Relatedworkstable} summarizes the aforementioned contributions with a comparison to this work. 

\begin{table*}[ht!]
\centering
\caption{Summary of related works}
 \label{Relatedworkstable}
 \resizebox{\textwidth}{!}{
\begin{tabular}
{|p{0.5cm}| p{5cm}| p{3.5cm}|  p{3.5cm}|  p{3.4cm}| p{3.4cm}|p{1.9cm}|}\hline

\textbf{Ref.} & \textbf{Contribution} & \textbf{Used Models}  & \textbf{Used Data} & \textbf{AD Tasks} & \textbf{Performance Metrics} & \textbf{Harsh Env.} \\\hline
\cite{chen2023driving} & Design and validation of an LLM Driver that interprets and reasons about driving situations to generate adequate actions.&  GPT-3.5; RL agent trained with PPO, lanGen, and trainable LoRA modules.  & 160k QA driving pairs dataset; Control Commands Dataset.  & Perception and action prediction. & Traffic light detection accuracy; Acceleration, brake, and steering errors. & No \\\hline
\cite{jin2023surrealdriver}  & LLM-based agent in a simulation framework to manage complex driving tasks. &  GPT-3 and GPT-4. & Driving Behavior Data; Simulation data from CARLA simulator; NLP libraries.  & Perception, control, and decision-making. & Collision rate.& No \\\hline
\cite{yang2024driving} &  LLM-based driver agents with reasoning and decision-making, aligned with human driving styles. &  GPT-4. & Private dataset from a human driver; NLP library; RL library.   & Behavioral alignment; Human-in-the-Loop system. & Collision rate, average speed, throttle percentage, and brake percentage. &  No \\\hline
\cite{guo2024co} & Co-driver agent, based on a vision language model, for planning, control, and trajectory prediction.  & Qwen-VL (9.6 billion parameters), including a visual encoder, a vision-language adapter, and the Qwen LLM. & CARLA simulation data; ROS2; Customized dataset. 

& Adjustable driving behaviors; Trajectory and lane prediction; Planning and control.  & Fluctuations frequency, and running time. & Foggy/gloomy; Rainy/gloomy   \\\hline
 \cite{wu2023language} & PromptTrack framework for 3D object detection and tracking by integrating cross-modal features within prompt reasoning.  &  VoVNetV2 to extract visual features; RoBERTa to embed language prompts. & NuPrompt for 3D perception in AD. &  3D object detection and tracking; Scene understanding.   & Average multiple object tracking precision (AMOTA), and identity switches (IDS).  &  No \\\hline
 \cite{ding2023hilm}  & High-resolution scene understanding using proposed HiLM-D. & BLiP-2; Q-Former; MiniGPT-4.   & DRAMA dataset; Pytorch; Visual Encoder; Query detection module; ST-adapter module.   & Risky object detection; Vehicle intentions' and motions' prediction.  & Mean intersection over union (mIoU) for detection; BLEU-4, METEOR, CIDER, and SPICE for captioning. & No   \\\hline
  
 \textbf{This work} & Implementation of harsh environment scenarios and performance evaluation of an MLLM-based driving agent.  & GPT-4o.  & Dedicated Prompt.  & Perception and decision-making. & Safety, comfort, efficiency, and speed scores. & Heavy rain; Storm; Foggy; Wetness; Good \\\hline
\end{tabular}
}
\end{table*}

\section{Background}
\begin{figure*}[t]
   \centering
  \includegraphics[width = 0.82\textwidth] {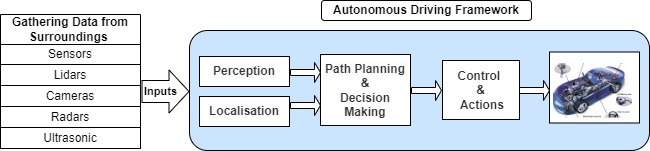}
  \caption{Typical architecture of ADS.}
  \label{adsframework}
\end{figure*}

The key modules associated with an AV are \textit{perception}, \textit{planning}, \textit{localization}, \textit{decision-making}, and \textit{action or control} as highlighted in Fig. \ref{adsframework}. The \textit{perception} module ensures sensing of the surroundings environment and identifies the driving scene. The main goal of the \textit{localization} module is to estimate with high precision and accuracy the vehicle position on the map. \textit{Path planning and decision} modules establish the waypoints that the vehicle should follow when moving through the surroundings. The set of waypoints corresponds to the vehicle trajectory. The \textit{action and control} module deals with the system's actuation such as braking, steering, and acceleration. 


Several MLLMs have been developed \cite{fourati2024xlm}, with a subset of them suitable for ADS as discussed in Section II. In this work, we opt for GPT-4o \cite{islam2024gpt}, an optimized version of GPT-4 built around the latter's foundational transformer architecture with enhancements in understanding, generating, and maintaining context across interactions. GPT-4o includes various specialized modules to improve its operation, namely (i) language understanding module, (ii) language generation module, (iii) context management module, (iv) task-specific modules, and (v) optimization and fine-tuning modules. Programming GPT-4o involves using APIs or SDKs, custom prompts, or fine-tuning and customization. In our work, interaction with GPT-4o is realized via API calls, where we utilize a customized prompt to guide the model response. In Table~\ref{comparison}, we provide a comparison of the most recent GPT models.

\begin{table*}[ht!]
\centering
\caption{Comparison of recent GPT models}
\label{comparison}
 \resizebox{\textwidth}{!}{
\begin{tabular}
{|p{3.2cm}|| p{3.2cm}| p{3.2cm}|  p{3.2 cm}|p{3.2 cm}|}\hline
       \textbf{Attribute} & \textbf{GPT-4o} & \textbf{GPT-4} & \textbf{GPT-4 Turbo} & \textbf{GPT-3.5} \\
        \hline \hline
        Initial Release Date & May 13, 2024 & March 14, 2023 & July 31, 2023 & March 15, 2022 \\
        \hline
        Modality & Audio, video, text \& images & Text \& images & Text \& images & Text \\
        \hline
        Context window & 128000 & 8192 & 25600 & 4096 \\
        \hline
        Parameters & Yet-to-Disclose (YTD) & 1.76 trillion & YTD & 175 billion \\
        \hline
        Cost per 1M token & Input: \$5, Output: \$15 & Input: \$30, Output: \$60 & Input: \$10, Output: \$20 & Input: \$1.5, Output: \$2 \\
        \hline
        Decoder layers & YTD & 120 & YTD & 96 \\
        \hline
        Prompting Method & YTD & Chain-of-Thought, n-shot & Chain-of-Thought, n-shot & n-shot \\
        \hline
        Training Data & YTD & 13T tokens & YTD & Over 570 GB of text data \\
        \hline
        Response Time & 65.8 ms per token & 94 ms per token & 50 ms per token & 35 ms per token \\
        \hline
        
        Type of Optimizer & YTD & AdamW & AdamW & Adam \\
        \hline
        Fine-Tuning Capability & Yes & Yes & Yes & Yes \\
        \hline
        Special Features & Multi-modal integration & Complex task handling & Optimized for performance and efficiency & Basic text understanding \\
        \hline
    \end{tabular}}
    
\end{table*}
\section{Proposed MLLM-based Driving Agent in a Closed-Loop Environment}
  In this section, we present the tools used to integrate the proposed MLLM-based driving agent, a.k.a., MLLM-AD-4o. 

\subsection{CARLA Simulator}
Car Learning to Act (CARLA) is an open-source simulator designed for AD research \cite{dosovitskiy2017carla}. It facilitates the development, training, prototyping, and validation of autonomous urban driving systems, including perception and control. CARLA offers open-source code and protocols, and a repository of digital assets (e.g., urban layouts, multitude of buildings, vehicles, pedestrians, street signs, etc.). The platform allows for customized sensor configurations and versatile environmental settings. CARLA can be used to evaluate the effectiveness of three distinct methods to ADS: (i) A traditional modular pipeline, including a vision-based perception module, a rule-based planner, and a maneuver controller;
(ii) A deep network that maps sensory inputs to driving commands, trained end-to-end via imitation learning; and (iii) an end-to-end model trained using RL.

CARLA has been built as an open-source layer over Unreal Engine 4 (UE4). It operates as a server-client system. Specifically, the server handles the simulation and scene rendering running on port 2000, while the client, implemented in Python, manages the interaction with the autonomous agent through socket communication. The client sends commands and meta-commands to the server and receives sensor data in return. 
In addition, the Python API serves as the primary client interface for CARLA enabling interaction with the simulation (including vehicle control), sensor data retrieval, and environment manipulation.
Also, CARLA provides diverse testing environments, e.g., 14 towns, each with unique layouts, road networks, and features. In our work, we focused on using Town-06, which is a low-density town that has long highways/roads with multiple lanes (4 to 6) per direction, several highway entrances and exits, and special junctions. 
\subsection{SUMO Simulator}
Simulation of Urban MObility (SUMO) is an open-source modular traffic simulation tool designed to handle large-scale traffic scenarios \cite{kusari2022enhancing}. It enables the simulation of vehicular traffic, pedestrian movements, and multimodal traffic flows across complex road networks. It offers detailed modeling of various traffic scenarios including vehicle behaviors and environments. It can simulate large-scale traffic systems involving autonomous and traditional vehicles in both urban and highway scenarios. SUMO is integrated with \textit{LimSim} as the engine managing traffic flows and vehicle behaviors, thus enabling \textit{LimSim} to simulate realistic and dynamic environments that reflect the complexity of urban traffic systems.
 
\subsection{LimSim++ Framework}
LimSim++ is a closed-loop framework for deploying MLLMs for autonomous driving, where multimodal data, such as text, audio, and video, can be analyzed, and then accurately react to driving scenarios \cite{fu2024limsim++}. LimSim++ integrates algorithms for real-time decision-making, obstacle detection, and navigation, making it an interesting tool for designing, developing, testing, and refining ADS.

\begin{figure*}
   \centering
   \includegraphics[width = 0.75\textwidth] {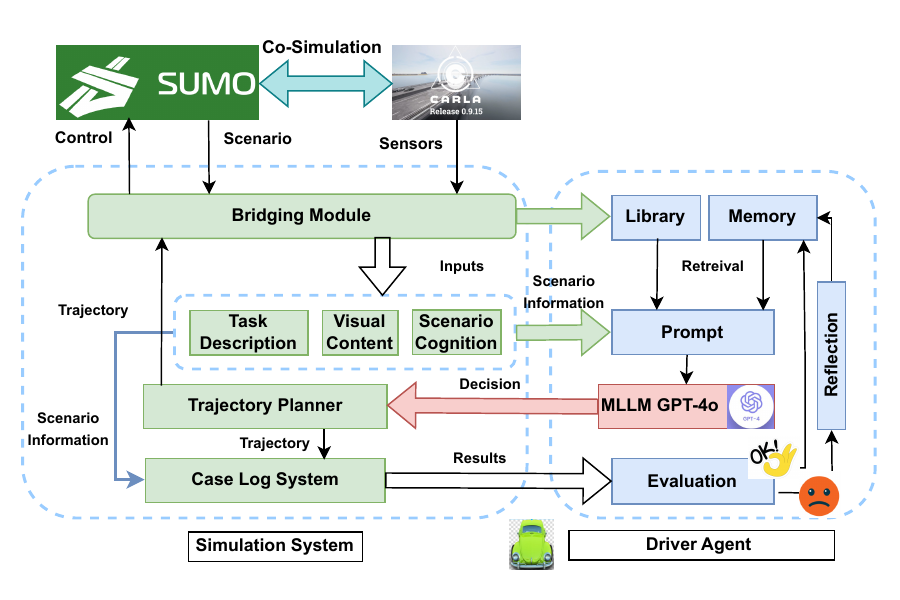}
  \caption{Integration of MLLM-AD-4o in LimSim++. }
   \label{limsim}
   \end{figure*}

To do so, LimSim++ simulates a closed-loop system with specifications of the environment including road topology, dynamic traffic flow, navigation, and traffic control. The MLLM's agent is based on prompts that provide real-time scenario information in visuals or text. The MLLM-supported driving agent system can process information, use tools, develop strategies, and assess itself. In Fig. \ref{limsim}, we depict the architecture of Limsim++ and explain its main modules as follows: 
\begin{itemize}
    \item \textit{Simulation system}: It gathers the scenario information from SUMO and CARLA, and packages it as input for the MLLM, such as visual content, scenario cognition, and task description. 
     \item \textit{Driver Agent}: It communicates with the ``simulation system'' to make driving judgments and employs MLLMs to interpret them. Data is represented as prompts for MLLMs to make suitable driving decisions. The outputs of MLLMs influence vehicle behavior and are kept in the case log system for knowledge accumulation. Following the simulation, the decisions are evaluated as follows: Those that performed well are immediately added to the memory, while poor decisions are added after reflection. The memory helps MLLMs make better driving decisions and improve the agent's performance.
\end{itemize}



When designing our MLLM-AD-4o, we customized the aforementioned modules as follows:
\begin{enumerate}
    \item We modified the original ``MPGUI.py'' file within LimSim++ and its related dependencies 
    to include the visualization of six cameras (rather than the default three front cameras) in the autonomous driving simulation. Hence, a comprehensive view is provided, which is crucial to evaluating the perception capability of the driving system, analyzing/debugging the AV’s sensor inputs, and decision-making. 
    
   \item We updated the default ``VLM-Driver-Agent'' to allow changes in the scenario and integrate new sensors and environmental conditions.

   \item We created a novel function in the CARLA simulator to automate the setup of different weather conditions.
\end{enumerate}



\section{Experimental Results}
\subsection{Definition of Metrics and Parameters}
To accurately assess the AD performance of the proposed MLLM-AD-4o, we introduce the following performance metrics (called scores):
\begin{itemize}
  \item \textit{Safety score:} The safety level is commonly assessed through Time-to-Conflict (TTC). When TTC falls below a specific threshold, a potential risk warrants penalties. The safety score can be given by
  \begin{equation}
    \label{EquatCollision score}
    \text{Safety\_score} = 
    \begin{cases}
    1, & \text{if } \tau_e \geq \tau_{\text{th}} \\
    \tau_e / \tau_{\text{th}}, & \text{otherwise,}
    \end{cases}
\end{equation}
where $\tau_e$ is the TTC of the AV and $\tau_{\text{th}}$ is the threshold value of TTC. Higher values reflect safer traveling.
    
\item \textit{Comfort score:} It evaluates the smoothness of the AV ride. Using empirical data, reference values can be determined for different driving styles, such as cautious, normal, and aggressive. It is computed as the average of the summation of acceleration score (acc\_score), jerk score, lateral acceleration score (lat\_acc\_score), and lateral jerk score (lat\_jerk\_score), 
as shown below    
    \begin{eqnarray}
    \label{EquaComfort}
    \text{Comfort\_score} &=& \frac{1}{4} \left( \text{acc\_score} + \text{jerk\_score}\right. \\ &+& \left. \text{lat\_acc\_score} + \text{lat\_jerk\_score}\right). \nonumber
\end{eqnarray}
A higher comfort score means that traveling is smoother.
    \item \textit{Efficiency score:} Driving efficiency is computed according to the vehicle's speed.
    In regular traffic conditions, the AV should maintain a speed at least equivalent to the average speed
    of surrounding vehicles. In sparse traffic conditions, the AV should approach the road's speed limit for efficiency. This score is computed as
    \ref{EquaEffeciency}.
    \begin{equation}
    \label{EquaEffeciency}
    \text{Efficiency\_score} = \begin{cases} 
    1.0, & \text{if } v_e \geq v^* \\
    \frac{v_e}{v^*}, & \text{otherwise,}
    \end{cases}
    \end{equation}
    where \( v_e \) represents the speed of the AV and \( v^* \in \{v_{avg}, v_{limit}\} \) is the targeted speed for efficiency, being \( v_{avg} \) as the average speed of surrounding vehicles, and \( v_{limit} \) as the road's speed limit. A high value means efficient traveling.
    
    \item \textit{Speed score:} The speed limit score (or speed score) penalizes the vehicle for exceeding the speed limit. This score is set to 0.9 when the AV exceeds the speed limit and 1 otherwise. This score is expressed as 
    \begin{equation}
    \label{EquaSpeedLimit}
    \text{Speed\_score} = 0.9^{\left(\frac{\text{Nbr\_frames\_with\_speeding}}{\text{Total\_nbr\_frames}} \times 10\right)},
    \end{equation}
   where ``Nbr\_frames\_with\_speeding'' is the number of captured frames while speeding, and ``Total\_nbr\_frames'' is the total number of captured frames.

\end{itemize}


In the following simulations, we set up $\alpha_1=\alpha_2=0.25$, $\alpha_3=0.5$, and $v_{limit}=50$ km/h (i.e., 13.89 m/s).

        
    \label{hyperparameter1}
\subsection{Interaction with GPT-4o API }
\begin{figure*}
    \centering
    \includegraphics[trim={0 1.5cm 0 0},clip,width = 0.99\textwidth] {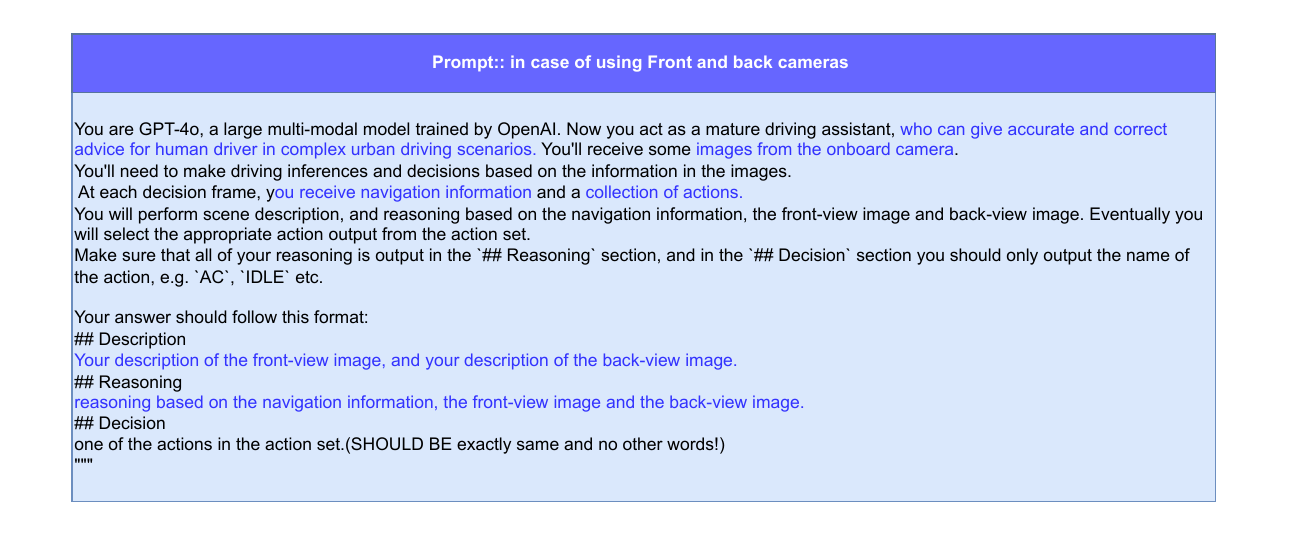}
    \caption{Example of used prompt for GPT-4o.}
    \label{promptexample}
\end{figure*}

To use the GPT-4o API with prompts to process the driving environment, with data mainly collected from CARLA and SUMO, 
we conduct the following steps \cite{GithubSonda}:
\begin{enumerate}
    \item Install OpenAI package and get OpenAI API key.
    \item Capture images from the front and back cameras (in CARLA), save them in a suitable format, and then convert them for API submission.
    \item Prepare the API request. It corresponds to sending the prompt with the image data to the GPT-4o API. The prompt describes what we want the model to do with the images and environmental data. In our work, we ask the model to analyze the surroundings, then, provide a driving decision. Fig.~\ref{promptexample} presents an example of used prompts for captured images with the AV's cameras.
    \item Process the response received from the API.
\end{enumerate}

\subsection{Setup of Weather Conditions}
In this work, we considered, in addition to ``Good weather'', four harsh weather conditions, namely ``Heavy rain'', ``Storm'', ``Foggy'', and ``Wetness''.
We developed a new function in the CARLA simulator, named ``set\_weather'', to configure different weather conditions. 
This function has eight parameters: 
\begin{itemize}
    \item \textit{Cloudiness:} The amount of cloud cover in the sky. 
    A high value means more clouds.
    \item \textit{Precipitation:} The intensity of rain. 
    A high value means heavier rain.
    \item \textit{Precipitation\_deposits:} The amount of water accumulating on the ground due to precipitation. 
    A high value means important accumulation.
    \item \textit{Wind\_intensity:} The strength of wind. 
    A high value means stronger wind.
    \item \textit{Sun\_altitude\_angle:} The angle of the sun above the horizon. Lower values can simulate dawn, dusk, or low-light conditions.
    \item \textit{Fog\_density:} The density of fog in the environment. A high value implies thicker fog.
    \item \textit{Fog\_distance:} The distance at which the fog starts to become noticeable.
    \item \textit{Wetness:} The wetness of the road surfaces. A high value makes the roads wetter.
\end{itemize}
In Table ~\ref{callweathersetting}, we summarize the parameters' values to set up each weather condition using the function ``set\_weather'' \cite{GithubSonda}.
 


\begin{table}[t]
\centering
\caption{Parameters to set up the weather conditions}
 \label{callweathersetting}
 \resizebox{0.48\textwidth}{!}{
\begin{tabular}
{|p{0.9cm}| p{7.9cm}| }\hline

\textbf{Case} & \textbf{Set\_weather parameters values} \\\hline
Heavy Rain & self.set\_weather(cloudiness=\textcolor{blue}{80.0}, precipitation=\textcolor{blue}{70.0}, precipitation\_deposits=\textcolor{blue}{60.0}, wind\_intensity=\textcolor{blue}{30.0}, sun\_altitude\_angle=\textcolor{blue}{45.0}, fog\_density=\textcolor{blue}{10.0}, fog\_distance=\textcolor{blue}{10.0}, wetness=\textcolor{blue}{80.0})  \\\hline
Storm & self.set\_weather(cloudiness=\textcolor{blue}{80.0}, precipitation=\textcolor{blue}{100.0}, precipitation\_deposits=\textcolor{blue}{100.0}, wind\_intensity=\textcolor{blue}{100.0}, sun\_altitude\_angle=\textcolor{blue}{20.0}, fog\_density=\textcolor{blue}{20.0}, fog\_distance=\textcolor{blue}{10.0}, wetness=\textcolor{blue}{80.0})  \\\hline
 Fog & self.set\_weather(cloudiness=\textcolor{blue}{40.0}, precipitation=\textcolor{blue}{5.0}, precipitation\_deposits=\textcolor{blue}{5.0}, wind\_intensity=\textcolor{blue}{10.0}, sun\_altitude\_angle=\textcolor{blue}{60.0}, fog\_density=\textcolor{blue}{70.0}, fog\_distance=\textcolor{blue}{3.0}, wetness=\textcolor{blue}{10.0}) \\\hline 
Wetness & self.set\_weather(cloudiness=\textcolor{blue}{30.0}, precipitation=\textcolor{blue}{0.0}, precipitation\_deposits=\textcolor{blue}{0.0}, wind\_intensity=\textcolor{blue}{0.0}, sun\_altitude\_angle=\textcolor{blue}{70.0}, fog\_density=\textcolor{blue}{0.0}, fog\_distance=\textcolor{blue}{0.0}, wetness=\textcolor{blue}{100.0}) \\\hline     
Good Weather & self.set\_weather(cloudiness=\textcolor{blue}{00.0}, precipitation=\textcolor{blue}{00.0}, precipitation\_deposits=\textcolor{blue}{00.0}, wind\_intensity=\textcolor{blue}{00.0}, sun\_altitude\_angle=\textcolor{blue}{60.0}, fog\_density={00.0}, fog\_distance=\textcolor{blue}{20.0}, wetness=\textcolor{blue}{00.0})  \\\hline 
\end{tabular}
}
\end{table}
{\subsection{Integration of Semantic LiDAR in LimSim++} CARLA supports various sensing tools, including traditional LiDAR and semantic LiDAR. Traditional LiDAR provides raw distance measurements in the form of a point cloud, with no inherent understanding of the objects in the scene. In contrast, semantic LiDAR provides both the point cloud and an understanding of what each point represents, thus simplifying tasks that involve scene understanding, object recognition, and data processing reduction. We opted here for semantic LiDAR. 
To add the latter in the perception environment within Limsim++, we follow several steps \cite{GithubSonda}: 
\begin{enumerate}
    \item Attach ``Semantic\_LIDAR'' function to the AV and gather data in CARLA.
    \item {Save and process LiDAR data}.
    \item Add ``Semantic\_LIDAR'' data to the prompt request.
\end{enumerate} }
\subsection{Simulation Results}

In this section, we assess the performances of the proposed MLLM-AD-4o autonomous driving agent in different conditions. Each experiment is built with the same scenario of our AV driving between points A and B, where it travels on multilane main roads and has to maneuver in curves. The autonomous driving agent has to take one decision among five possible decisions in each time frame, including idle (no change in current behavior), acceleration, deceleration, turn left, or turn right.    



\begin{figure*}
    \centering
    \begin{subfigure}{1\columnwidth}
        \centering
        \includegraphics[trim={0 0 0 0.5cm},clip,width=\textwidth]{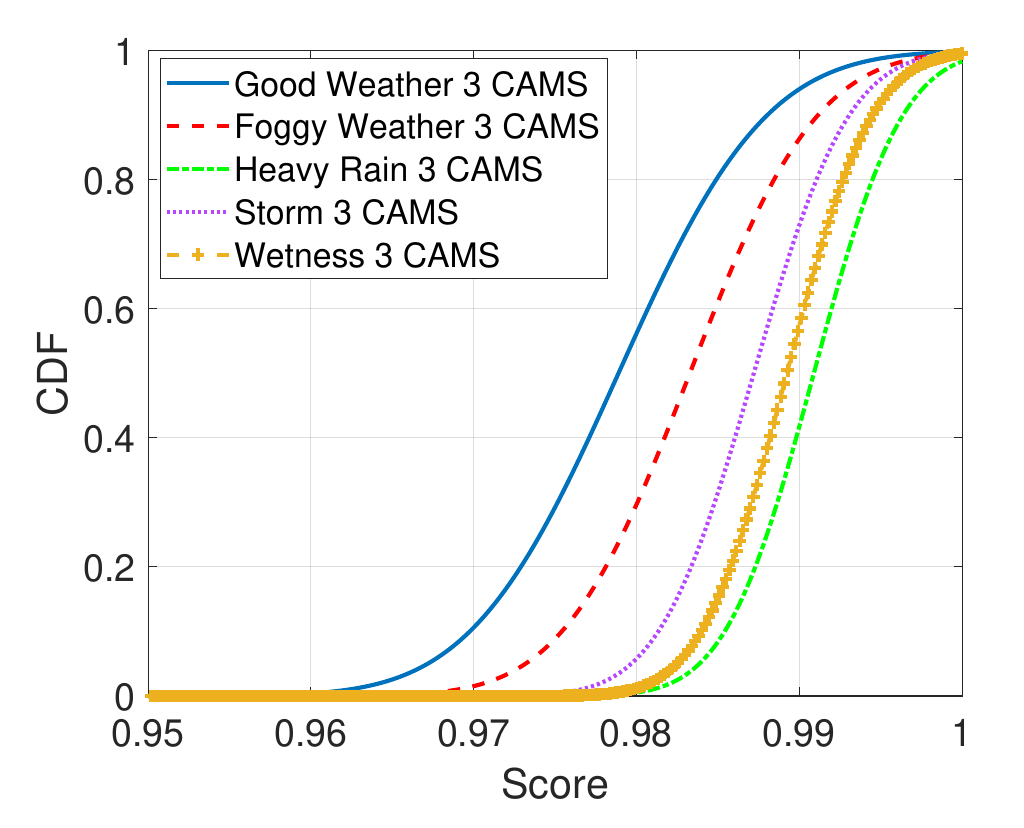}
        \subcaption{CDF of safety score}
        \label{collisionScore3cams}
    \end{subfigure}
    \hfill
    \begin{subfigure}{1\columnwidth}
        \centering
        \includegraphics[width=0.97\textwidth]{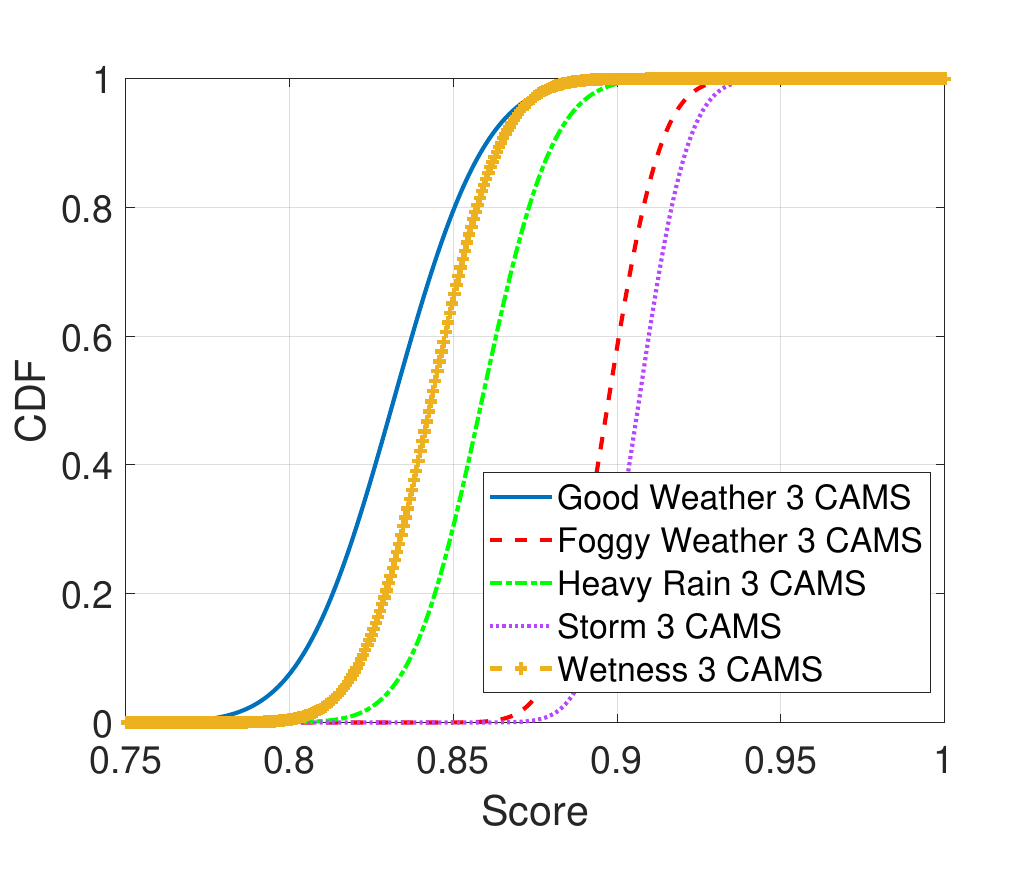}
        \subcaption{CDF of comfort score}
        \label{ComfortScore3cam}
    \end{subfigure}
    \hfill
    \begin{subfigure}{1\columnwidth}
        \centering
        \includegraphics[trim={0 0 1.1cm 1.1cm},clip,width=\textwidth]{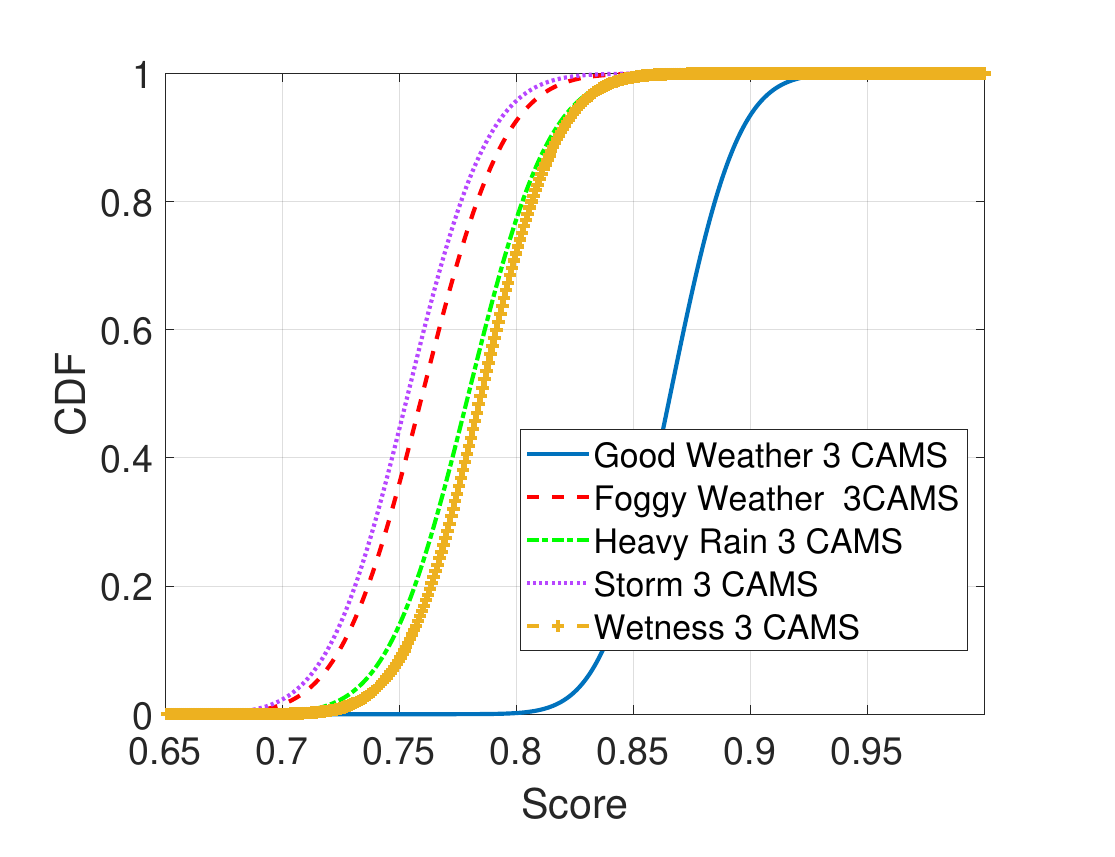}
        \subcaption{CDF of efficiency score}
        \label{EffeciencyScore3cam}
    \end{subfigure}
    \hfill
    \begin{subfigure}{1\columnwidth}
        \centering
        \includegraphics[width=0.95\textwidth]{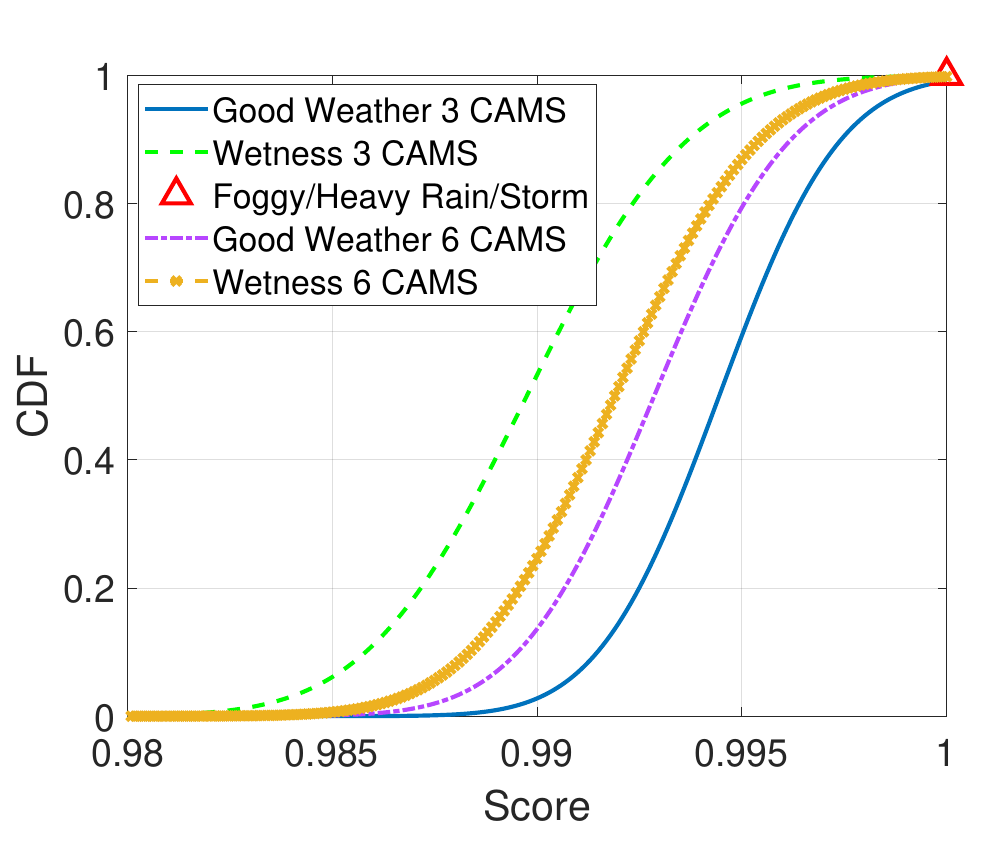}
        \subcaption{CDF of speed score}
        \label{SpeedScore3cam}
    \end{subfigure}
    \caption{CDFs of MLLM-AD-4o scores (Front cameras).}
    \label{Scores3cam}
\end{figure*}
Figs.~\ref{Scores3cam}a-\ref{Scores3cam}d present the cumulative distribution functions (CDFs) for the safety, comfort, efficiency, and speed scores/metrics, respectively, under various weather conditions, when 3 cameras are used for autonomous driving by the proposed MLLM-AD-4o method\footnote{Fig. \ref{Scores3cam}d illustrates also speed score results for the 6 cameras case, which will be discussed later.}. 
In Fig. \ref{Scores3cam}a, we see that when the weather is good (blue line), the agent takes a riskier behavior that may cause collisions than in a harsh condition. This also impacts the passengers' comfort during travel as shown in Fig. \ref{Scores3cam}b. Nevertheless, in good weather, the driving behavior is the most efficient one as shown in Fig. \ref{Scores3cam}c and it aligns with the speed performance depicted in Fig. \ref{Scores3cam}d, 
This trade-off highlights that while the AV may be driven more aggressively and with less comfort, its efficiency improves as it aligns more closely with the surrounding traffic's speed.
In contrast, in harsh weather conditions, e.g., heavy rain (green line) or wetness (dark yellow line), safety is enforced through slower driving, which translates into a better comfort score, offering a smoother and more cautious ride, at the expense of degraded efficiency and speed performances.

Among the weather conditions, we notice in Fig. \ref{Scores3cam} that the best safety CDF is achieved in heavy rain (green line), the best comfort CDF in stormy weather (pink line), the best efficiency CDF in good weather, and the best speed CDF in all of the foggy, heavy rain, and stormy weather conditions (red triangle). 
This suggests that the MLLM-AD-4o driving agent considers heavy rain the most unsafe environment requiring low-speed traveling. Also, traveling can be smooth in stormy weather, while driving is most efficient in good weather. Finally, in foggy/heavy rain/storm conditions, the agent is more likely to respect the speed limits as it favors slower driving to guarantee security, which is not the case in good and wet weather conditions. These results reflect the trade-offs the MLLM-AD-4o agent makes between various performance scores, under different weather patterns.

\begin{figure*}
    \centering
    \begin{subfigure}{0.32\textwidth}
        \centering
        \includegraphics[trim={0 0  0.5cm 0},clip,width=\textwidth]{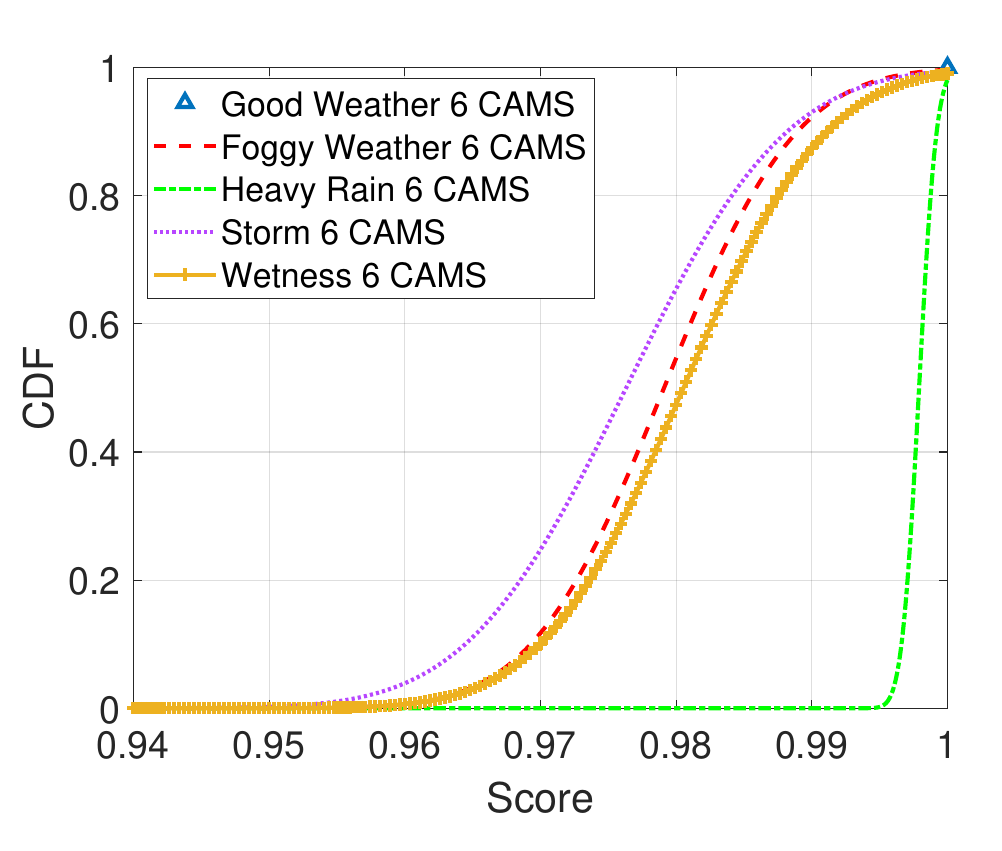}
        \subcaption{CDF of safety score}
        \label{collisionScore6cams}
    \end{subfigure}
    \hfill
    \begin{subfigure}{0.32\textwidth}
        \centering
        \includegraphics[trim={0 0  0.5cm 0},clip,width=\textwidth]{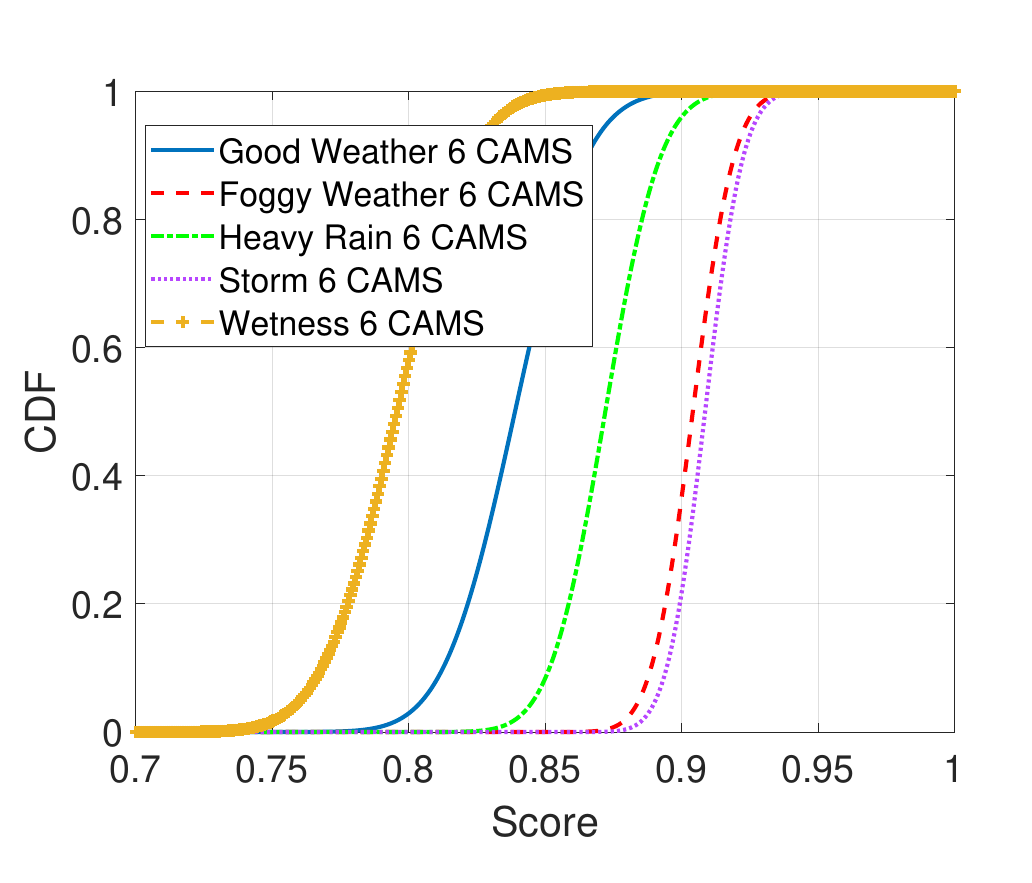}
        \subcaption{CDF of comfort score}
        \label{ComfortScore6cam}
    \end{subfigure}
    \hfill
    \begin{subfigure}{0.32\textwidth}
        \centering
        \includegraphics[trim={0 0  0.5cm 0},clip,width=\textwidth]{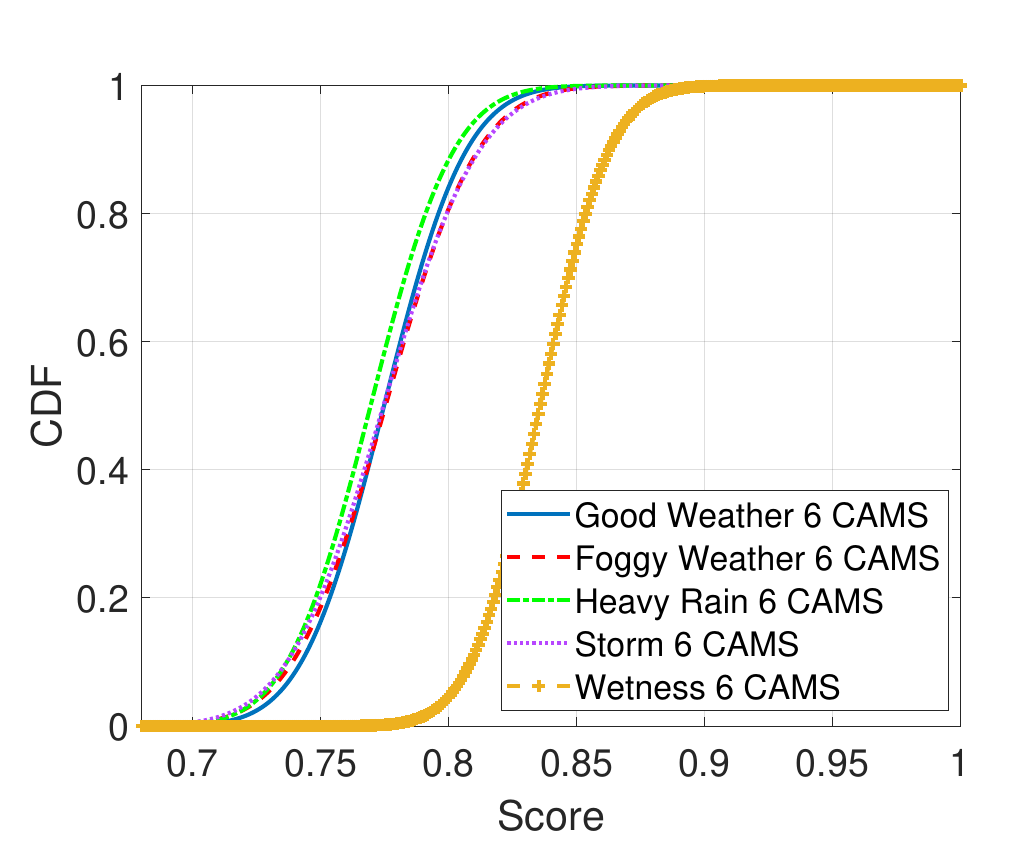}
        \subcaption{CDF of efficiency score}
        \label{EffeciencyScore6cam}
    \end{subfigure}
    \caption{CDFs of MLLM-AD-4o scores (Front+rear cameras).}
    \label{Scores6cam}
\end{figure*}
In Figs. \ref{Scores6cam}a-\ref{Scores6cam}c, we depict the CDFs of the safety, comfort, and efficiency scores when 6 cameras (3 front + 3 rear) are used by the AD agent. In good weather, MLLM-AD-4o achieves now the best safety CDF performance (blue triangle), at the expense of comfort and efficiency. Indeed, the additional information brought by the 3 rear cameras makes the agent more precautious regarding the behavior of the surrounding vehicles, which was not possible when it only relied on 3 cameras. This is confirmed with the speed CDF result (purple line) in Fig. \ref{Scores3cam}d which is higher than that of the 3 cameras' CDF. According to Fig. \ref{Scores6cam}b, the best CDFs are realized in stormy/foggy weather (with a preference for stormy) at the expense of efficiency and safety. Indeed, in these situations, the AD agent is more likely to drive smoothly with minor behavior changes in response to poor visibility, with a higher safety risk and reduced efficiency, due to the misalignment with the surrounding road traffic behavior. Moreover, the best efficiency CDF is obtained in wet conditions, as shown in Fig. \ref{Scores6cam}c. Indeed, with the knowledge of its 6 cameras, the AV maintains a speed closer to the average of other vehicles by continuously adapting to the dynamics of the road and surrounding traffic. This is validated through the results in Fig. \ref{Scores3cam}d where the 6 cameras' speed CDF is better than the 3 cameras' one in wetness. However, as illustrated in Figs. \ref{Scores6cam}a-\ref{Scores6cam}b, this behavior negatively impacts the safety and comfort scores.
In summary, the addition of the rear cameras enhances the vehicle's ability to perceive and respond to potential collision threats, particularly in challenging weather conditions.

\begin{figure*}
    \centering
        \begin{subfigure}{0.32\textwidth}
        \centering
        \includegraphics[width=\textwidth]{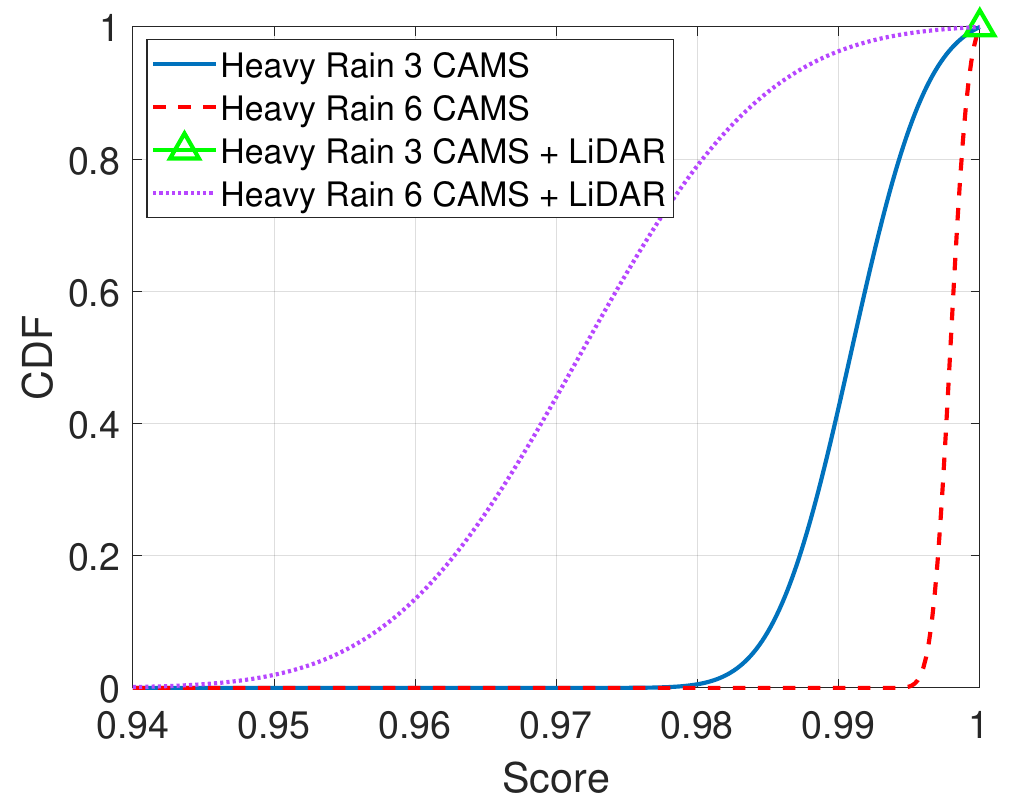}
        \subcaption{CDF of safety score}
        \label{collisionScoreLidarHeavyRain}
    \end{subfigure}
        \begin{subfigure}{0.32\textwidth}
        \centering
        \includegraphics[width=0.98\textwidth]{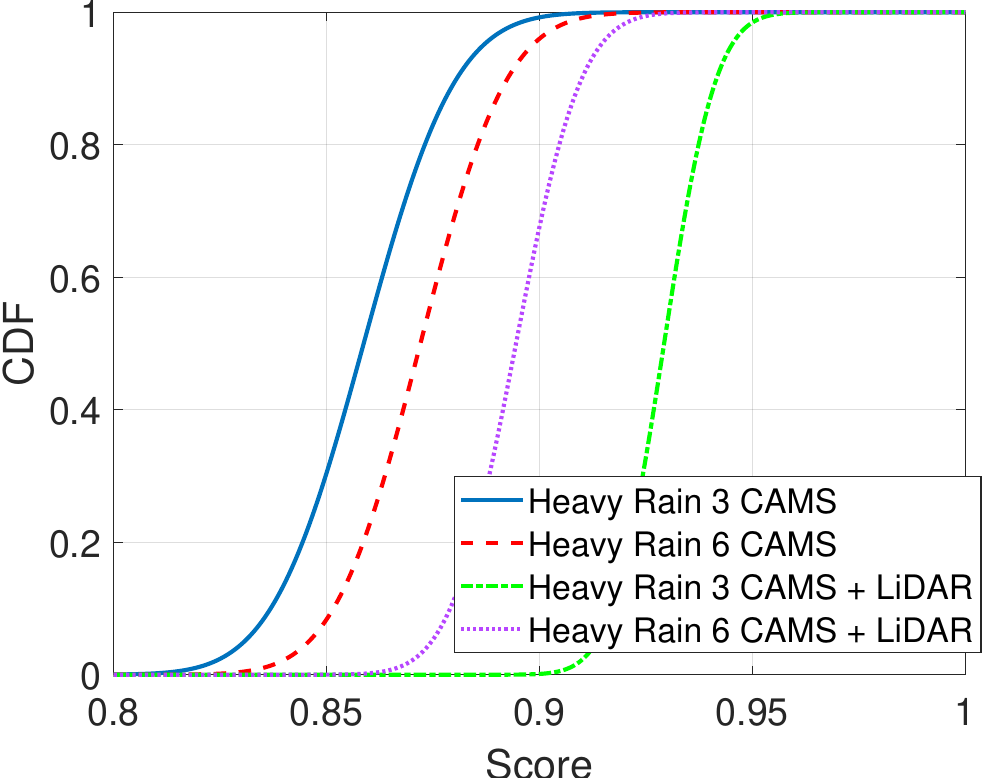}
        \subcaption{CDF of comfort score}
        \label{ComfortScoreLidarHeavyRain}
    \end{subfigure}   
    \begin{subfigure}{0.32\textwidth}
        \centering
        \includegraphics[width=0.97\textwidth]{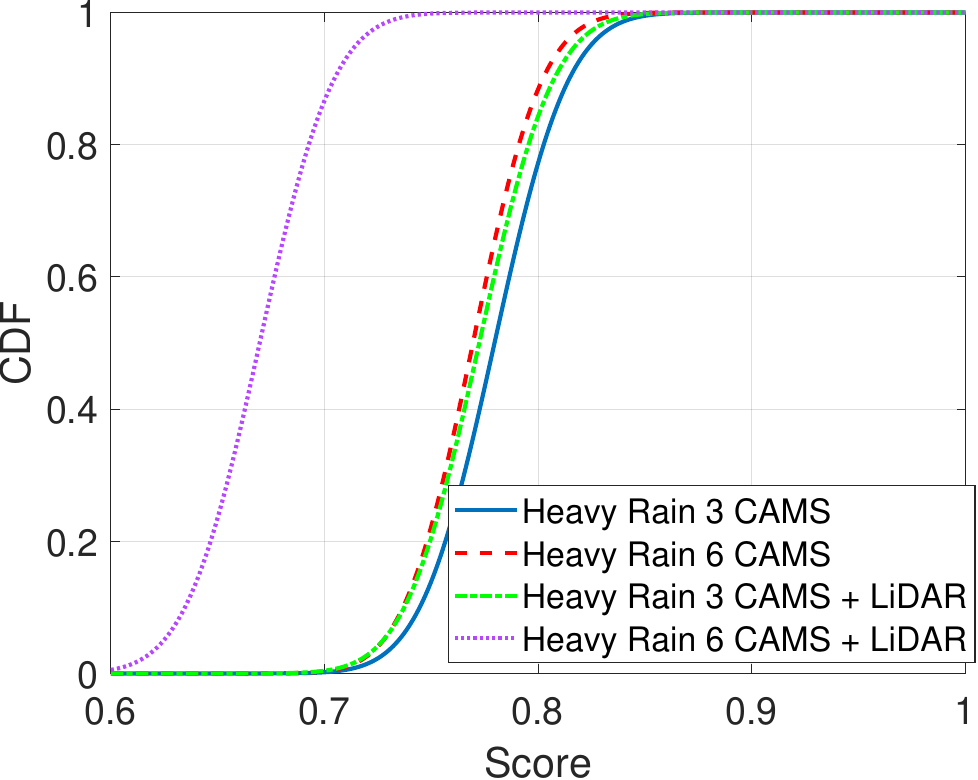}
        \subcaption{CDF of efficiency score}
        \label{EfficiencyScoreLidarHeavyRain}
    \end{subfigure}
    \caption{CDFs of MLLM-AD-4o scores (Combinations of cameras with LiDAR; Heavy rain).}
    \label{ScoresLidar}
\end{figure*}

Figs. \ref{ScoresLidar}a-c illustrate the CDFs of the safety, comfort, and efficiency scores, when the AD agent is equipped with 3 (front) cameras only, 6 (3 front + 3 rear) cameras only, 3 (front) cameras + LiDAR, and 6 (3 front + 3 rear) cameras + LiDAR, respectively, operating in a heavy rain condition. The best safety CDF is obtained when the LiDAR is used with 3 cameras (green triangle in Fig. \ref{ScoresLidar}a). Indeed, activating the LiDAR in adverse weather conditions enhances the system's awareness of its environment leading to more responsible driving actions, and thus to an improved safety. 
Nevertheless, safety and efficiency CDFs degrade when LiDAR is activated with 6 cameras. This suggests that rear cameras' information might distort the agent's comprehension of its surroundings when combined with LiDAR. The best comfort score is also achieved when 3 cameras are activated with LiDAR (green line in Fig. \ref{ScoresLidar}b), while the best efficiency score is realized with a 3-camera system (blue line in Fig. \ref{ScoresLidar}c). Nevertheless, the activation of LiDAR with the 3 cameras (green line in Fig. \ref{ScoresLidar}c) realizes similar efficiency CDF performance to the latter. Based on these results, the activation of LiDAR with cameras should be triggered in a timely manner, in particular conditions, and with specific configurations, as the full combination of cameras with LiDAR does not necessarily lead to better AD performances.

\section{Conclusion}
Following integrating the harsh environment conditions into the CARLA/Limsim++ simulation framework, we proposed a novel AD agent based on the GPT-4o MLLM model for driving perception and control decision-making, a.k.a., MLLM-AD-4o. Through extensive experiments, we assessed the performances of MLLM-AD-4o in terms of safety, comfort, efficiency, and speed score CDFs, for different weather conditions, and given diverse combinations of cameras and LiDAR activated at the AV.      
Our findings indicate that activating both front and rear cameras significantly enhances the system's performance, compared to the case where only the front cameras are used, especially in good weather conditions. Moreover, activating LiDAR with only 3 cameras provides the best performance compared to the cases where only cameras are used. However, when LiDAR is used with all 6 cameras, the AV's performance degrades, suggesting that the right combination of cameras and LiDAR should be activated in specific configurations and weather conditions. 
The obtained results draw important insights into the efficient design of AD systems, where careful data fusion should be accurately realized to optimize autonomous driving behavior.




\section*{Acknowledgment}
This work is funded by a Mitacs Globalink Research Award scholarship.

\bibliographystyle{IEEEtran}
\balance
\bibliography{ref.bib}

\begin{thebibliography}{10}
\providecommand{\url}[1]{#1}
\csname url@samestyle\endcsname
\providecommand{\newblock}{\relax}
\providecommand{\bibinfo}[2]{#2}
\providecommand{\BIBentrySTDinterwordspacing}{\spaceskip=0pt\relax}
\providecommand{\BIBentryALTinterwordstretchfactor}{4}
\providecommand{\BIBentryALTinterwordspacing}{\spaceskip=\fontdimen2\font plus
\BIBentryALTinterwordstretchfactor\fontdimen3\font minus \fontdimen4\font\relax}
\providecommand{\BIBforeignlanguage}[2]{{%
\expandafter\ifx\csname l@#1\endcsname\relax
\typeout{** WARNING: IEEEtran.bst: No hyphenation pattern has been}%
\typeout{** loaded for the language `#1'. Using the pattern for}%
\typeout{** the default language instead.}%
\else
\language=\csname l@#1\endcsname
\fi
#2}}
\providecommand{\BIBdecl}{\relax}
\BIBdecl

\bibitem{chang2024survey}
Y.~Chang, X.~Wang, J.~Wang, Y.~Wu, L.~Yang, K.~Zhu, H.~Chen, X.~Yi, C.~Wang, Y.~Wang \emph{et~al.}, ``A survey on evaluation of large language models,'' \emph{ACM Trans. Intelli. Syst. Technol.}, vol.~15, no.~3, pp. 1--45, 2024.

\bibitem{fourati2024xlm}
S.~Fourati, W.~Jaafar, N.~Baccar, and S.~Alfattani, ``{XLM} for autonomous driving systems: A comprehensive review,'' \emph{arXiv preprint arXiv:2409.10484}, 2024.

\bibitem{wang2024exploring}
Y.~Wang, W.~Chen, X.~Han, X.~Lin, H.~Zhao, Y.~Liu, B.~Zhai, J.~Yuan, Q.~You, and H.~Yang, ``Exploring the reasoning abilities of multimodal large language models ({MLLM}s): A comprehensive survey on emerging trends in multimodal reasoning,'' \emph{arXiv preprint arXiv:2401.06805}, 2024.

\bibitem{cui2024survey}
C.~Cui, Y.~Ma, X.~Cao, W.~Ye, Y.~Zhou, K.~Liang, J.~Chen, J.~Lu, Z.~Yang, K.-D. Liao \emph{et~al.}, ``A survey on multimodal large language models for autonomous driving,'' in \emph{Proceed. of the IEEE/CVF Winter Conf. Appl. Comput. Vis.}, 2024, pp. 958--979.

\bibitem{wen2023dilu}
L.~Wen, D.~Fu, X.~Li, X.~Cai, T.~Ma, P.~Cai, M.~Dou, B.~Shi, L.~He, and Y.~Qiao, ``Dilu: A knowledge-driven approach to autonomous driving with large language models,'' \emph{arXiv preprint arXiv:2309.16292}, 2023.

\bibitem{ding2023hilm}
X.~Ding, J.~Han, H.~Xu, W.~Zhang, and X.~Li, ``{HiLM-D}: Towards high-resolution understanding in multimodal large language models for autonomous driving,'' \emph{arXiv preprint arXiv:2309.05186}, 2023.

\bibitem{yuan2024rag}
J.~Yuan, S.~Sun, D.~Omeiza, B.~Zhao, P.~Newman, L.~Kunze, and M.~Gadd, ``{RAG-Driver}: Generalisable driving explanations with retrieval-augmented in-context learning in multi-modal large language model,'' in \emph{Proc. Robotics: Science and Syst. (RSS) Conf.}, 2024, pp. 1--14.

\bibitem{wang2024drivecot}
T.~Wang, E.~Xie, R.~Chu, Z.~Li, and P.~Luo, ``Drivecot: Integrating chain-of-thought reasoning with end-to-end driving,'' \emph{arXiv preprint arXiv:2403.16996}, 2024.

\bibitem{wang2023drivemlm}
W.~Wang, J.~Xie, C.~Hu, H.~Zou, J.~Fan, W.~Tong, Y.~Wen, S.~Wu, H.~Deng, Z.~Li \emph{et~al.}, ``{DriveMLM}: Aligning multi-modal large language models with behavioral planning states for autonomous driving,'' \emph{arXiv preprint arXiv:2312.09245}, 2023.

\bibitem{huangdrivlme}
Y.~Huang, J.~Sansom, Z.~Ma, F.~Gervits, and J.~Chai, ``{DriVLMe}: Exploring foundation models as autonomous driving agents that perceive, communicate, and navigate,'' in \emph{Proc. Vis. and Langua. for Autonom. Driv. and Roboti. Wrkshp.}, 2024.

\bibitem{liao2024vlm2scene}
G.~Liao, J.~Li, and X.~Ye, ``{VLM2Scene}: Self-supervised image-text-{LiDAR} learning with foundation models for autonomous driving scene understanding,'' in \emph{Proc. AAAI Conf. on Artifi. Intelli.}, vol.~38, no.~4, 2024, pp. 3351--3359.

\bibitem{wang2024omnidrive}
S.~Wang, Z.~Yu, X.~Jiang, S.~Lan, M.~Shi, N.~Chang, J.~Kautz, Y.~Li, and J.~M. Alvarez, ``{OmniDrive}: A holistic {LLM}-agent framework for autonomous driving with {3D} perception, reasoning and planning,'' \emph{arXiv preprint arXiv:2405.01533}, 2024.

\bibitem{tian2024enhancing}
H.~Tian, K.~Reddy, Y.~Feng, M.~Quddus, Y.~Demiris, and P.~Angeloudis, ``Enhancing autonomous vehicle training with language model integration and critical scenario generation,'' \emph{arXiv preprint arXiv:2404.08570}, 2024.

\bibitem{mao2023language}
J.~Mao, J.~Ye, Y.~Qian, M.~Pavone, and Y.~Wang, ``A language agent for autonomous driving,'' in \emph{Proc. Conf. Langua. Model. (COMS)}, 2024.

\bibitem{rayfeedback}
J.~Z. Z. H.~A. Ray and E.~Ohn-Bar, ``Feedback-guided autonomous driving,'' 2024.

\bibitem{ashqar2024leveraging}
H.~I. Ashqar, T.~I. Alhadidi, M.~Elhenawy, and N.~O. Khanfar, ``Leveraging multimodal large language models ({MLLMs}) for enhanced object detection and scene understanding in thermal images for autonomous driving systems,'' \emph{Automat.}, vol.~5, no.~4, pp. 508--526, 2024.

\bibitem{luo2024delving}
S.~Luo, W.~Chen, W.~Tian, R.~Liu, L.~Hou, X.~Zhang, H.~Shen, R.~Wu, S.~Geng, Y.~Zhou \emph{et~al.}, ``Delving into multi-modal multi-task foundation models for road scene understanding: From learning paradigm perspectives,'' \emph{IEEE Trans. Intelli. Veh. (Accepted)}, 2024.

\bibitem{li2024dense}
R.~Li, Z.~Zhang, C.~He, Z.~Ma, V.~M. Patel, and L.~Zhang, ``Dense multimodal alignment for open-vocabulary {3D} scene understanding,'' \emph{arXiv preprint arXiv:2407.09781}, 2024.

\bibitem{xu2023drivegpt4}
Z.~Xu, Y.~Zhang, E.~Xie, Z.~Zhao, Y.~Guo, K.-Y.~K. Wong, Z.~Li, and H.~Zhao, ``{DriveGPT4}: Interpretable end-to-end autonomous driving via large language model,'' \emph{IEEE Robot. Automat. Lett.}, vol.~9, no.~10, pp. 8186--8193, Oct. 2024.

\bibitem{cui2024receive}
C.~Cui, Y.~Ma, X.~Cao, W.~Ye, and Z.~Wang, ``Receive, reason, and react: Drive as you say, with large language models in autonomous vehicles,'' \emph{IEEE Intelli. Transport. Syst. Mag.}, 2024.

\bibitem{li2024unifiedmllm}
Z.~Li, W.~Wang, Y.~Cai, X.~Qi, P.~Wang, D.~Zhang, H.~Song, B.~Jiang, Z.~Huang, and T.~Wang, ``{UnifiedMLLM}: Enabling unified representation for multi-modal multi-tasks with large language model,'' \emph{arXiv preprint arXiv:2408.02503}, 2024.

\bibitem{GithubSonda}
``{MLLM} applied to autonomous driving across various weather conditions,'' \url{https://github.com/SondaFourati/MLLM-applied-to-autonomous-driving-across-various-weather-conditions/tree/main}, accessed: 2024-11-10.

\bibitem{islam2024gpt}
R.~Islam and O.~M. Moushi, ``{GPT}-4o: The cutting-edge advancement in multimodal {LLM},'' \emph{Authorea Preprints}, 2024.

\bibitem{chen2023driving}
L.~Chen, O.~Sinavski, J.~Hünermann, A.~Karnsund, A.~J. Willmott, D.~Birch, D.~Maund, and J.~Shotton, ``Driving with {LLM}s: Fusing object-level vector modality for explainable autonomous driving,'' in \emph{Proc. IEEE Int. Conf. Robot. Automat. (ICRA)}, May 2024, pp. 14\,093--14\,100.

\bibitem{jin2023surrealdriver}
Y.~Jin, X.~Shen, H.~Peng, X.~Liu, J.~Qin, J.~Li, J.~Xie, P.~Gao, G.~Zhou, and J.~Gong, ``Surrealdriver: Designing generative driver agent simulation framework in urban contexts based on large language model,'' \emph{arXiv preprint arXiv:2309.13193}, 2023.

\bibitem{yang2024driving}
R.~Yang, X.~Zhang, A.~Fernandez-Laaksonen, X.~Ding, and J.~Gong, ``Driving style alignment for llm-powered driver agent,'' \emph{arXiv preprint arXiv:2403.11368}, 2024.

\bibitem{guo2024co}
Z.~Guo, A.~Lykov, Z.~Yagudin, M.~Konenkov, and D.~Tsetserukou, ``Co-driver: {VLM}-based autonomous driving assistant with human-like behavior and understanding for complex road scenes,'' \emph{arXiv preprint arXiv:2405.05885}, 2024.

\bibitem{wu2023language}
D.~Wu, W.~Han, T.~Wang, Y.~Liu, X.~Zhang, and J.~Shen, ``Language prompt for autonomous driving,'' \emph{arXiv preprint arXiv:2309.04379}, 2023.

\bibitem{dosovitskiy2017carla}
A.~Dosovitskiy, G.~Ros, F.~Codevilla, A.~Lopez, and V.~Koltun, ``{CARLA}: An open urban driving simulator,'' in \emph{Proc. Conf. Robot Learn.}\hskip 1em plus 0.5em minus 0.4em\relax PMLR, 2017, pp. 1--16.

\bibitem{kusari2022enhancing}
A.~Kusari, P.~Li, H.~Yang, N.~Punshi, M.~Rasulis, S.~Bogard, and D.~J. LeBlanc, ``Enhancing {SUMO} simulator for simulation based testing and validation of autonomous vehicles,'' in \emph{Proc. IEEE Intelli. Veh. Symp. (IV)}.\hskip 1em plus 0.5em minus 0.4em\relax IEEE, 2022, pp. 829--835.

\bibitem{fu2024limsim++}
D.~Fu, W.~Lei, L.~Wen, P.~Cai, S.~Mao, M.~Dou, B.~Shi, and Y.~Qiao, ``{LimSim++}: A closed-loop platform for deploying multimodal {LLMs} in autonomous driving,'' in \emph{Proc. IEEE Intelli. Veh. Symp. (IV)}, 2024, pp. 1084--1090.

\end{thebibliography}

\end{document}